\documentclass[letterpaper]{article} 
\usepackage{aaai24}  
\usepackage{times}  
\usepackage{helvet}  
\usepackage{courier}  
\usepackage[hyphens]{url}  
\usepackage{graphicx} 

\usepackage{multirow}
\usepackage{booktabs,subcaption,amsfonts,dcolumn}
\usepackage{amsmath}
\usepackage{amssymb}

\usepackage{natbib}  
\usepackage{caption} 
\usepackage{algorithm}
\usepackage{algorithmic}

\usepackage{newfloat}
\usepackage{listings}
\DeclareCaptionStyle{ruled}{labelfont=normalfont,labelsep=colon,strut=off} 
\floatstyle{ruled}
\newfloat{listing}{tb}{lst}{}
\floatname{listing}{Listing}
\title{Reinforcement Learning Based Sensor Optimization for Bio-markers}
\author{
    Sajal Khandelwal \quad 
    Pawan Kumar \quad 
    Syed Azeemuddin \quad 
    Zia Abbas
}
\affiliations{
    \textsuperscript{\rm 1}International Institute of Information Technology\\

    C R Rao Road, Gachibowli\\
    Hyderabad, 500032, India\\
    pawan.kumar@iiit.ac.in
}

\usepackage{bibentry}

\begin{document}

\maketitle

\begin{abstract}
Radio frequency (RF) biosensors, in particular those based on inter-digitated capacitors (IDCs), are pivotal in areas like biomedical diagnosis, remote sensing, and wireless communication. Despite their advantages of low cost and easy fabrication, their sensitivity can be hindered by design imperfections, environmental factors, and circuit noise. This paper investigates enhancing the sensitivity of IDC-based RF sensors using novel reinforcement learning based Binary Particle Swarm Optimization (RLBPSO), and it is compared to Ant Colony Optimization (ACO), and other state-of-the-art methods. By focusing on optimizing design parameters like electrode design and finger width, the proposed study found notable improvements in sensor sensitivity. The proposed RLBPSO method shows best optimized design for various frequency ranges when compared to current state-of-the-art methods. 
\end{abstract}

\section{Introduction}

Radio frequency (RF) bio-sensors have become a crucial component in various fields, including biomedical diagnosis, remote sensing, and wireless communication. Among different types of RF sensors, inter-digitated capacitors (IDCs) have emerged as a popular choice due to their low cost, simple design, and ease of fabrication. However, the sensitivity of IDC-based RF sensors can be limited by various factors, such as non-idealities in device design, environmental interference, and circuit noise.

Several methods have been proposed to overcome these limitations and improve the sensitivity of IDC-based RF sensors, including the use of heuristic optimization algorithms such as Binary Particle Swarm Optimization (BPSO) \cite{modBpso}, Ant Colony Optimization (ACO) \cite{ACOar}, Artificial Bee Colony (ABC) \cite{ABCar}, Simulated Annealing (SA) \cite{SAar}, Ant Lion Optimization (ALO) \cite{ALOar}, and Reinforcement Learning based Binary Particle Swarm Optimization(RLBPSO) \cite{RLBPSOar}. These algorithms can analyze and optimize parameters affecting IDC-based RF bio-sensor sensitivity, such as the inter-digitated electrode design, the number of fingers, finger width, finger gap, and the electrode material.

This research paper presents a comprehensive study on enhancing the sensitivity of IDC-based RF sensors using heuristic optimization algorithms with modification. The parameters that impact the sensitivity of the sensor are analyzed, and these algorithms are applied to identify the optimal design parameters. In this work, the performance of the optimized IDC-based RF sensor is compared with the work published in \cite{KUNALar}, and a significant improvement in sensitivity is demonstrated.

The proposed approach has the potential to advance the development of highly sensitive RF sensors, enabling their use in diverse applications such as environmental monitoring, medical diagnostics, and wireless communication. The optimized sensor design may pave the way for new and innovative applications by improving the sensitivity, accuracy and reliability of RF sensors.

\section{Methodology}
\subsection{Initial IDC Design}
The initial design of the IDC (Inter-Digitated Capacitor) was inspired by the work done in \cite{KUNALar} but with modifications in dimensions. The IDC sensor was fabricated on a 1.6 mm FR-4 substrate measuring $94.5 \times 36$ mm2. An epoxy-resin-based mask measuring $87 \times 36$ mm2 was deposited on the IDC side. A sample chamber made of grey PLA material with dimensions of $27 \times 19.5$ mm2 and a cavity of $24 \times 16.5$ mm2 was 3D printed. To achieve impedance matching, the width of the microstrip line was set at 3.13 mm to obtain an input characteristic impedance of 50 ohms.

The figure shows the IDC-based RF sensor used for simulation and how the inter-digSTATEd region is divided into a cell pattern of $16 \times 11$ cells, each with a size of 1.5 mm. A specific design structure is created by combining k cells, where $k$ is the total number of unknown parameters (96) in the optimization problem for IDC. The cell pattern of IDC is kept in an anti-symmetric structure across the horizontal axis, similar to the conventional IDC structure. IDC behaves like a band-pass filter, and the S11 parameter is focused on observing the sensor's resonant frequency change.

\subsection{Cost Function}
Designing an efficient IDC RF-based biosensor involves several parameters that must be optimised to achieve optimal performance. To this end, we propose a methodology for optimising an IDC-based biosensor using a heuristic optimisation algorithm. The methodology aims to improve the sensitivity of the biosensor and enhance its accuracy in detecting analytes. 
The sensitivity of the biosensor is directly related to the shift in resonating frequency(${\Delta}F$), which increases with a more significant resonating shift and decreases with a lower one. 
\[{\Delta}f = {f_{\text{sam}} - f_{\text{ref}}}. \]
The normalised frequency (NF) of the sensor is \cite{KUNALar}:
\[ NF = \frac{{\Delta}f}{f_{\text{ref}}}.\]
Since the sensitivity is directly proportional to the normalised frequency of the sensor.
So, in this work, the inverse of NF is taken as the optimisation algorithm's objective function(OF). Now, optimisation algorithms aim to find the minima of the OF. Now, The OF is defined as: 
\begin{equation}{\label{OF}}
    OF = \frac{f_{\text{ref}}}{{\Delta}f} = \frac{f_{\text{ref}}}{f_{\text{sam}} - f_{\text{ref}}},
\end{equation}
where ${f_{\text{ref}}}$ and ${f_{\text{sam}}}$ are the resonating frequency of the sensor when reference and sample are placed on the sensor, respectively.

\subsection{Optimisation Algorithms}
The biosensor sensitivity optimization process depends on multiple factors. So, this is a complex problem to solve. Heuristic optimization algorithms are commonly used for complex and large-scale problem spaces because they offer efficient and practical approaches to finding better or near-optimal solutions. In this work, we cover the popular heuristic algorithms, which are well-proven and cover the latest optimization in this area. 

\subsubsection{Binary Particle Swarm Optimisation}\cite{modBpso}
The Particle Swarm Optimization (PSO) \cite{PSOar} algorithm is a powerful metaheuristic optimization technique that draws inspiration from the social behaviour of birds and insects. This study uses a Binary Particle Swarm Optimization (BPSO), with a modification update in the particle's position in PSO, as mentioned in \cite{modBpso}. It searches for the optimal cell pattern that maximizes the fitness value and minimizes the cost value. The optimization process involves initializing the swarm of particles with random positions and velocities within the search space, evaluating the cost of each particle in the swarm, updating personal and global best positions and cost, updating velocities and positions of the particles, and repeating the process until the convergence criteria met. The optimal cell pattern of the IDC sensor is obtained from the global best position of the swarm.

In BPSO, finding the best solution involves a balance between exploration and exploitation. The swarm size directly affects the exploration and exploitation process, as mentioned in \cite{BPSO1ar}. The swarm size is taken as 25 in this study, which is near optimum as mentioned in \cite{BPSO1ar}. The BPSO equations are
\begin{align}{\label{bpsoeq1}}
     {V_{n}}^{t+1} &= w{V_{n}}^{t} + c_{1}r_{1}({P_{n}}^{t} - {X_{n}}^{t}) \\
     &+ c_{2}r_{2}(G^{t} - {X_{n}}^{t}), \nonumber 
\end{align}
where $V_{n}$, $X_{n}$ and $P_{n}$ are the velocity, position and personal best of the $n^{th}$ particle, respectively, $G$ is the global best solution of the whole swarm,  $w$ is called the inertia weight, $c_1$ and $c_2$ are personal and social acceleration coefficient. Here, $V$, $X$, $P$, and $G$ are $k$-dimensional vectors. The values of $w$, $a_1$, $a_2$ and $k$ are $1$, $2$, $2$, and $96$ respectively. The values of $w$, $c_1$ and $c_2$ are selected for the better optimal solution as mentioned in \cite{modBpso}.

\begin{equation}{\label{bpsoeq2}}
    {\tt TF}({V_{n}}^{t+1}) = 
        \begin{cases}
            \frac{2}{{1+\exp(-a{V_{n}}^{t+1})}} - 1, & {V_{n}}^{t+1} > 0 \\
            1 - \frac{2}{{1+\exp(-a{V_{n}}^{t+1})}}, & {V_{n}}^{t+1} \leq 0,
        \end{cases}
\end{equation}
where $a$ in \ref{bpsoeq2} is defined as
\begin{equation}{\label{bpsoeq3}}
    a = e - ((e-d)/i),
\end{equation}
where ${\tt TF}({V_{n}}^{t+1})$ is the transfer function of the velocity of $n^{th}$ particle at $(t + 1)$ iterations, as mentioned in \cite{modBpso}, $a$ is an iterative parameter called transfer factor, $i$ is the number of iterations, $e$ and $d$ are the maximum and minimum transfer factors, respectively. The values of $e$ and $d$ are $2$ and $1$, respectively, for early convergence of the algorithm. The position of the $n^{th}$ particle at the $(t + 2)$ iteration is given by
\begin{align}{\label{bpsoeq4}}
    {X_{n}}^{t+2} = 
        \begin{cases}
            1, & {\tt TF}({V_{n}}^{t+1}) > r,  \: {X_{n}}^{t+1} = 0 \\
            0, & {\tt TF}({V_{n}}^{t+1}) > r,  \: {X_{n}}^{t+1} = 1 \\
            {X_{n}}^{t+1}, & {\tt TF}({V_{n}}^{t+1}) \leq r. \\
        \end{cases}
\end{align}

The simulation results from Table \ref{algoComparison} show that the BPSO algorithm effectively optimizes the sensitivity of the IDC sensor by optimizing the cell pattern. The optimal cell pattern obtained from the algorithm maximizes the sensitivity of the IDC sensor by maximizing the normalized frequency. The proposed approach can be used for designing and optimizing IDC sensors for various applications.

In conclusion, the BPSO algorithm is an effective optimization technique that can be used to optimize the sensitivity of IDC sensors. The proposed approach is effective in optimizing the cell pattern of the IDC sensor, and the results demonstrate the effectiveness of the algorithm in maximizing the sensitivity of the IDC sensor. The proposed approach can be used for designing and optimizing IDC sensors for various applications.




\begin{algorithm}
    \caption{BPSO Optimization}
    \label{BPSO_code}
    \begin{algorithmic}[1]

        \STATE Initialize population of particles with random cell patterns
        \STATE Calculate cost for each solution
        \STATE Initialize velocity for each particle
        \STATE Initialize global best solution and its cost

        
        \FOR {$i=1,2,\ldots, {\tt maxIter}$}
            \FOR {${\tt eachParticle}$}
                \STATE Update velocity \textbf{eq. [needs to be added]}
                \STATE Update particle position \textbf{eq. [needs to be added]}
                \STATE Calculate Cost of new position
                \IF{${\tt cost{\_}new} < {\tt particleBestCost}$}
                    \STATE Update Particle best cost and position
                \ENDIF
                \IF{${\tt cost{\_}new} < {\tt globalBestCost}$}
                    \STATE Update global best cost and position
                \ENDIF
            \ENDFOR

            \STATE Calculate the cost for each solution
            \STATE Update the global best solution
        \ENDFOR
        \STATE \textbf{Return} global best solution
    \end{algorithmic} 
\end{algorithm}

\subsubsection{Artificial Bee Colony Algorithm}
Artificial Bee Colony (ABC) optimization \cite{ABCar} is a metaheuristic algorithm based on the foraging behaviour of honeybees. The algorithm mimics the food source exploration of bees and uses a population of artificial bees to search for the optimal solution in a given search space. The ABC algorithm works by initializing a population of artificial bees and assigning them to search for food sources (candidate solutions). Three types of bees are used in the algorithm: employed bees, onlookers bees and scouts bees. Each food source represents a possible solution to the optimization problem: a combination of cell patterns for the IDC sensor.

During the search process, the bees can exploit the information contained in the previously found food sources and explore new ones using local and global search strategies. The global search is performed by scout bees that randomly generate new solutions, while the local search is performed by employed bees that exploit the information contained in the food sources already found.

The ABC algorithm iteratively performs the search process until a termination criterion is met. At the end of the search process, the best solution found is returned as the optimal cell pattern for the IDC sensor.

\begin{algorithm}[ht]
    \caption{ABC Optimization}
    \label{ABC_code}
    \begin{algorithmic}[1]

        \STATE Set colony size, $N = 50$ 
        \STATE Set maximum number of ${\tt iterations}, {\tt maxIter} = 50$ 
        \STATE Set limit for number of consecutive failures, limit = 50 
        \STATE Initialize population of bees with random cell patterns
        \STATE Calculate the cost for each solution
        
        \FOR {${\tt iteration}=1,2,\ldots, {\tt maxIter}$}
            \FOR {${\tt eachEmployedBees}$}
                \STATE Select neighbour solution
                \STATE Compute cost of the neighbour solution
                \IF {${\tt cost{\_}neighbour} < {\tt employeeBeeBestCost}$} 
                    \STATE Update employed bee solution to neighbour
                \ELSE 
                    \STATE Increase employed bee trial's count
                \ENDIF
            \ENDFOR
            
            \FOR {${\tt eachOnlookerBees}$}
                \STATE Select a neighbour solution based on the probability of its cost
                \STATE Compute cost of the neighbour solution
                \IF {${\tt cost{\_}neighbour} < {\tt onLookerBeeBestCost}$} 
                    \STATE Update onlooker bee solution to neighbour
                \ELSE 
                    \STATE Increase onlooker bee trial count
                \ENDIF
            \ENDFOR

            \FOR {${\tt eachScoutBeeswithTrialCount} < {\tt limit}$}
                \STATE Select a neighbour solution based on the probability of its cost
                \STATE Generate a random solution
                \STATE Reset the scout bee's trial count
            \ENDFOR

            \STATE Calculate the cost for each solution
            \STATE Update the global best solution
        \ENDFOR
        \STATE \textbf{Return} global best solution
    \end{algorithmic} 
\end{algorithm}

\subsubsection{Ant Colony Optimisation.}
Ant Colony Optimization (ACO) \cite{ACOar} is another meta-heuristic algorithm inspired by the foraging behaviour of ants. The ACO algorithm works by initializing a population of artificial ants and assigning them to search for a food source (candidate solution). Each food source represents a possible solution to the optimization problem, which is a combination of cell patterns for the IDC sensor.
During the search process, the ants can exploit the information in the previously found food sources and explore new ones using local and global search strategies. The global search is performed by pheromone trails, which represent the information about the quality of food sources found so far. The ants are more likely to choose food sources with higher pheromone concentrations, which indicates better solutions. The local search is performed by the ants' ability to evaluate the quality of the cell pattern and adjust it accordingly.

The ACO algorithm updates the quality of each food source (candidate solution) by using the fitness function. The algorithm also employs the concept of pheromones to encourage the exploration of new solutions. The amount of pheromone deposited on a food source is proportional to its quality, and the ants are more likely to choose food sources with high pheromone concentrations.

The ACO algorithm iteratively performs the search process until a termination criterion is met, such as a maximum number of iterations or reaching a predefined fitness value. At the end of the search process, the best solution found is returned as the optimal cell pattern for the IDC sensor.

\begin{algorithm}[ht]
    \caption{Ant Colony Optimization} 
    \label{ACO_code}
    \begin{algorithmic}[1]

        \STATE Initialize population of ants with random cell patterns
        \STATE Calculate cost for each solution
        \STATE Initialize pheromone matrix
        \STATE Initialize global best solution and its cost

        
        \FOR {${\tt iteration}=1,2,\ldots, {\tt maxIter}$}
            \FOR {${\tt eachAnt}$}
                \STATE Construct a solution using pheromone trail and heuristic information
                \STATE Calculate Cost of new position
                \IF{${\tt cost{\_}new} < {\tt globalBestCost}$}
                    \STATE Update Global best solution
                \ENDIF
                \STATE Update pheromone trail using \textbf{eq. [needs to be added]}
            \ENDFOR

            \STATE Update pheromone trail using elitist strategy
        \ENDFOR
        \STATE \textbf{Return} global best solution
    \end{algorithmic} 
\end{algorithm}

\subsubsection{Simulated Annealing.}
Simulated Annealing (SA) Optimisation \cite{SAar} is a probabilistic optimization algorithm inspired by the annealing process in metallurgy. The SA algorithm works by initializing a random solution (a cell pattern for the IDC sensor) and evaluating its fitness value (the normalized frequency). The algorithm then randomly perturbs the current solution and evaluates the fitness value of the new solution. If the new solution has a better fitness value, it is accepted as the current solution. However, if the new solution has a worse fitness value, it may still be accepted with a certain probability that decreases over time.
The acceptance probability is controlled by a parameter called the temperature, which decreases over time according to a cooling schedule. The cooling schedule determines the rate at which the temperature decreases and is crucial for the success of the algorithm. At high temperatures, the algorithm is allowed to accept solutions with worse fitness values, which can help the algorithm escape from local optima. As the temperature decreases, the algorithm becomes more selective and only accepts solutions that improve the fitness value.
The SA algorithm iteratively performs the perturbation and acceptance process until a termination criterion is met, such as a maximum number of iterations or reaching a predefined fitness value. At the end of the search process, the best solution found is returned as the optimal cell pattern for the IDC sensor.

\begin{algorithm}[ht]
    \caption{Simulated Annealing} 
    \label{SA_code}
    \begin{algorithmic}[1]

        \STATE Initialize a random pattern for IDC cells
        \STATE Calculate cost for initial solution
        \STATE Set initial and final temperature
        \STATE Set cooling rate

        
        \FOR {${\tt iteration}=1,2,\ldots, {\tt maxIter}$}
            \FOR {${\tt each particle}$}
                \STATE Generate a new solution by randomly changing cell patterns
                \STATE Calculate Cost of new position
                \IF{${\tt cost{\_}new} < {\tt currentBestCost}$}
                    \STATE Update current best solution
                \ELSE
                    \STATE Update new solution with probability $e^{-\delta/T}$
                \ENDIF
            \ENDFOR

        \ENDFOR
        \STATE \textbf{Return} global best solution
    \end{algorithmic} 
\end{algorithm}

\subsubsection{Ant Lion Optimization.}
The Ant Lion Optimisation (ALO) \cite{ALOar} algorithm is inspired by the interaction between antlions and ants, where antlions construct traps to capture ants, and ants exhibit gradient-based movement to avoid falling into the traps. This interaction is adapted into an optimization algorithm that efficiently explores the search space.

The ALO algorithm initializes a population of antlions and ants, with each individual representing a candidate solution (cell pattern) for the IDC sensor. Antlions update their positions based on the fitness values of the current solutions. Stronger antlions, corresponding to higher fitness values, construct traps in regions likely to contain better solutions.
Ants move within the search space using a gradient-based strategy, adjusting their positions towards regions with higher fitness values. This movement encourages the exploration of potential solutions. The fitness function evaluates each ant's cell pattern's fitness (normalized frequency). This step provides the reinforcement signal for the subsequent updates. Ants perform local searches in the vicinity of their positions, further exploring the search space. This local search enables the exploitation of regions with high fitness, enhancing the algorithm's convergence capabilities. Based on their local search and the positions of the traps, ants update their positions to avoid falling into traps and navigate towards high-fitness regions.

\begin{algorithm}[ht]
    \caption{ALO} 
    \label{ALO_code}
    \begin{algorithmic}[1]

        \STATE Initialize the positions of antlions
        \STATE Calculate the cost values of antlions
        \STATE Save the best antlion and its position (elite antlion)
        \FOR {${\tt iteration}=1,2,\ldots, {\tt maxIter}$}
            \FOR {${\tt each}$ ${\tt antlion}$}

                \STATE Select antlion using the roulette wheel method
                \STATE Slide randomly walking ants in a trap
                
                \STATE Generate ant's random walk route around elite antlion
                \STATE Generate the ant's random walk route around the selected antlion
                \STATE Normalize random walks
                \STATE Calculate the position of the ant
            \ENDFOR
            \STATE Calculate the cost values of ants
            \STATE Combine ants and antlions
            \STATE Sort according to their costs and take the first population size
            \STATE Update the elite antlion
        \ENDFOR

        \STATE \textbf{Return} best antlion
    \end{algorithmic} 
\end{algorithm}

\subsubsection{Reinforcement Learning based Binary Particle Swarm Optimization.}
The Reinforcement Learning-based Binary Particle Swarm Optimization (RLBPSO) algorithm is an advanced hybrid optimization method that combines the strengths of reinforcement learning and the Binary Particle Swarm Optimization (BPSO) algorithm. This method is used to guide the adaption of parameters of BPSO in a way that improves the overall performance of the algorithm. The flow of this algorithm involves states, action space, reward function, learning algorithm, and online adaption process. 
The state is defined by three parameters which are iteration percentage, diversity function and the no improvement iteration, where iteration percentage means the percentage of the iterations during the execution of the BPSO, and it ranges from 0-1, diversity function can be defined as the average of the Euclidean distances between each particle and the best particle and the no improvement iteration means the stagnant growth duration input, as stated by Equations 7, 8, 9 and 10 in \cite{RLBPSOar}. 

The action or the output of the action network is used to control the BPSO parameters $w, c_1$ and $c_2$, referenced from Equations 11 and 12 in \cite{RLBPSOar}.

The reward function for the RL agent is used to measure the reward value after the execution of an action to find the better global optimum. The reward function is mentioned in Equation 13 of \cite{RLBPSOar}.

This technique employs two neural networks to enhance the performance of the optimization process: the actor-network and the action-value network.


The actor network is specifically trained to assist the particles in the PSO in selecting optimal parameters for their current states. It takes as input three components: the proportion of iterations, the percentage of iterations without improvements, and the diversity of the swarm. Based on this information, the actor-network divides all particles into several groups, each with its action. In PSO, actions typically control parameters such as $w, c_1,$ and $c_2,$ but they can control any parameter if necessary.

The action-value network, on the other hand, is responsible for evaluating the performance of the actor network. It provides gradients for training the actor network and helps to fine-tune its performance. A reward function is used to train both networks. The design of this function is straightforward and aims to motivate the PSO to produce better solutions with each iteration.
The velocity update equation, after calculating the parameters from the action network, is
\begin{align} {\label{vel_RLBPSO}}
    V(t+1)_{i}^{d} &= w \times V(t)_{i}^{d} \\
    &+ c_1 \times r_1 \times ({\tt pbest}_{fi(d)}^{d} - X_{i}^{d}) \\ 
    &+ c_2 \times r_2 \times ({\tt gbest}_{i}^{d} - X_{i}^{d}) \\ 
    &+ c_3 \times r_3 \times ({\tt pbest}_{i}^{d} - X_{i}^{d}).
\end{align} 


\begin{algorithm}[ht]
    \caption{RLBPSO} 
    \label{RLBPSO_code}
    \begin{algorithmic}[1]

        \STATE Initialize the particle swarm and parameters
        \STATE Initialize particle velocity and position
        \STATE $n=25$ is number of swarms
        \WHILE {$f_e \leq f_{\max}$}
            \STATE Calculate {$s_t$} 
            \STATE Calculate {$a_t$} with actor-network $\mu$ and $s_t$
            \FOR {$k=1,2,\ldots, n$}
                \STATE Convert actions into operating parameters ($c_1, c_2, c_3$ and $c_4$)
                \STATE Calculate the updated velocity from the actions (\ref{vel_RLBPSO})
                \IF{${\tt rand}() < c_4 \times 0.01 \times {\tt flag}$} 
                    \STATE Reinitialize particle's position
                \ELSE 
                    \STATE Update position $x^k_{t+1} = x^k_t + v^k_t$
                \ENDIF
                \STATE Calculate the evaluation value of all particles
                \STATE Update the parameters in particle operation
            \ENDFOR
        \ENDWHILE
    \end{algorithmic} 
\end{algorithm}

\begin{table}[h]
    \begin{center}
    \caption{Best cost optimization algorithms comparison }
    \label{algoComparison}
    \begin{subtable}\linewidth
    \centering
        \caption{For 1.5 GHz IDC design}
        \begin{tabular}{|l|l|}
            \hline
            \textbf{Algorithms} & \textbf{Cost}\\ 
            \hline
                 BPSO & 27.314 \\
            \hline
                 SA with Random Mutation & 50.04 \\
            \hline
                 SA with Swap Mutation & 72.00 \\
            \hline
                 ABC & 30.89 \\            
            \hline
                 ACO & 80.76 \\
            \hline
                \textbf{ALO} & \textbf{21.30} \\
            \hline
                \textbf{RLBPSO} & \textbf{20.47} \\
            \hline
        \end{tabular}
    \end{subtable}\par
    \vspace{0.5cm}
    \begin{subtable}\linewidth
    \centering
        \caption{For 5Ghz IDC design}
        \begin{tabular}{|l|l|}
            \hline
            \textbf{Algorithms} & \textbf{Cost}\\ 
            \hline
                 BPSO & 26.13 \\
            \hline
                 SA with Random Mutation & 60.05 \\
            \hline
                 SA with Swap Mutation & 75.88 \\
            \hline
                 ABC & 26.36 \\            
            \hline
                 ACO & 45.31 \\
            \hline
                \textbf{ALO} & \textbf{23.44} \\
            \hline
                \textbf{RLBPSO} & \textbf{18.53} \\
            \hline
        \end{tabular}
    \end{subtable}
    \end{center}
\end{table}

\section{Numerical Experiments}
\subsection{Experimental Setup}
In this work, we are using CST studio suite software to find
the sensitivity of the biosensor. The input to this software is
a 96-size binary array, which states the design of the biosen-
sor to the software. In optimizing the biosensor’s design, we
use the inverse of biosensor sensitivity as a cost function for
the optimization algorithm. Our optimization goal is to in-
crease biosensor sensitivity, so algorithms aim to find the
minimum cost value. Assuming that the time CST software
takes to find the cost is f (independent of the time taken by
algorithms) and The time complexity of finding the cost of
one design pattern is O(f ) for all algorithms.

\subsection{Comparisons}
In Table \ref{algoComparison} (a), we compare the costs of the proposed method RLBPSO with other methods for 1.5 GHz IDC design and in Table \ref{algoComparison} (b), we show for 5GHz IDC design. We find that the proposed method gives the lowest cost among all the methods compared. The second best is achieved by ALO followed by BPSO. Hence, the reinforcement learning approach improves on BPSO method. For SA, random mutations gave better results compared to SA with swap mutation.

In Table \ref{tab:compare}, we compare the time for the proposed method RLBPSO with other methods. We show times for three runs. We observe that SA annealing takes the least time, followed by BPSO and RLBPSO. Although, SA is fastest, the cost achieved by SA is very high as seen in Table \ref{algoComparison} for both 1.5GHz and 5GHz IDC design. Hence, there is a trade-off for better sensor design at the cost of higher run time. 

In Table \ref{tab:Algorithms_Hyperparameters}, we show the optimal hyperparameters used for experiments. 

\begin{table}[!h]
    \centering
     \caption{Time Comparisons. Time shown in seconds.}
    \label{tab:compare}
\begin{tabular}{|l|lll|l|}
\hline
\multirow{2}{*}{Algorithms} & \multicolumn{3}{c|}{Time}\\
\cline{2-4}
& First & Second & Third \\
\hline
BPSO & 161930.73 & 165993.54 & 169367.05 \\
ACO & 169824.50 & 171153.61 & 167495.39 \\
ALO & 165763.77 & 157216.60 & 162813.72 \\
ABC & 197216.60 & 181413.07 & 179314.83  \\
RLBPSO & 163333.17 & 178683.38 & 171666.58 \\
SA & 65877.62 & 62680.29 & 63073.45 \\
\hline
\end{tabular}
\end{table}

\begin{table}[!h]
    \centering
    \caption{Hyperparameters}
    \label{tab:Algorithms_Hyperparameters}
    \begin{tabular}{|p{2cm}| p{4cm}|}
        \hline
        \textbf{Algorithms} & \textbf{Hyperparameters}\\ 
        \hline
        \textbf{BPSO} & Maximum Iterations = 25, Swarm size = 25 \\ 
        \hline
        \textbf{ACO} & Maximum Iterations = 25, Ants Pop. = 25 \\ 
        \hline
        \textbf{ALO} & Maximum Iterations = 25, Antlion Agents = 25 \\ 
        \hline
        \textbf{ABC} & Maximum Iterations = 25, Total bees = 30, Employed bees = 22, Onlooker bees = 5, Scout bees = 3\% \\ 
        \hline
        \textbf{RLBPSO} & Maximum Iterations = 25, Swarm size = 25 \\
        \hline
        \textbf{SA} & Maximum Iterations = 100 \\
        \hline
    \end{tabular}
\end{table}

\section{Complexity analysis}
In the following, we show complexity analysis of the methods compared. 

\subsection{BPSO} 
The time complexity of the Binary Particle Swarm Optimization (BPSO) algorithm can vary depending on several factors, including the number of particles $N$, the dimensionality of problem space $D$, and the termination criteria.
In BPSO, a population of particles explores the search space to find the optimal solution. Each particle maintains its position and velocity and updates them based on its own experience and the experience of neighbouring particles. The algorithm iteratively performs these updates until a termination condition is met.

In BPSO, The particles are randomly initialized in the search space. The time and space complexity of initialization is $O(ND)$, where $N$ is the number of particles, and $D$ is the problem's dimensionality. For each particle, the fitness function is evaluated to determine its quality. So, the time complexity of fitness evaluation will be $O(Nf)$. In each iteration, the particles update their positions and velocities based on their current positions, velocities, and the best positions. The time complexity of updating the positions and velocities is $O(N D)$, and for updating the particle's best solution and the global best solution is $O(N)$. 
The algorithm continues iterating until a termination condition is met, which could be a maximum number of iterations (${\tt maxItr}$) or reaching a desired fitness value. The time complexity of checking the termination condition is typically $O({\tt maxItr}).$

The total time taken by PSO algorithm is $O({\tt maxIt} (O(ND) + O(Nf) + O(N) + O(N)))$. So, the overall time complexity is $O({\tt maxIt} (N (D + f + 2))$ and space complexity is $O(ND)$. 

\subsection{ABC} 
In the Artificial Bee Colony (ABC) algorithm, Initialize the populations of employed, onlooker and scout bees in the search space, which takes $O((EB+OB+SB) D)$ space and time complexity, where $EB$, $OB$ and $SB$ are the total numbers of employed, onlooker and scout bees and $D$ is the space dimensions. For every iteration step, Each employed bee explores a solution within its neighbourhood, calculates its cost function, and shares all information with onlooker bees. Onlooker bees select employer bees' location based on the cost using probabilistic algorithms, perform some local search around the employed bees' solution, and evaluate the cost function of these solutions. And then, the scout bees identify solutions that have yet to improve for a certain number of iterations and randomly generate new solutions. So, the overall time taken by every iteration step is $O((EB + OB + SB)  f)$. Moreover, the maximum number of iterations ${\tt maxItr}$ is used as a termination criterion.
The total time complexity is $O({\tt maxIt} ((EB + OB + SB)  (f + D)))$ and space complexity is $O((EB + OB + SB)  D)$.

\subsection{ACO}
In Ant Colony Optimization (ACO) algorithm, the ants' population and the pheromone trail matrix are randomly initialized in the search space; both the time and space will take $O(ND)$ for the ants' population and $O(D^2)$ for the pheromone trail matrix, where $N$ is the total number of ants. $D$ is the dimensionality of the search space. For every iteration, the ants construct solutions by iteratively choosing the next component based on the pheromone trails and heuristic information and calculating its cost function, which takes $O(Nf)$.
After, all ants complete finding their solutions, the pheromone trails are updated based on their cost function value. Moreover, using the graph traversal method, the pheromone matrix is updated in $O(N + D)$ complexity. The total number of iterations used is ${\tt maxIt}$ as termination criteria.
So, the overall time taken by ACO is $O({\tt maxIt} (Nf + N + D))$. The time complexity is $O({\tt maxIt}(N(f + 1) + D))$ and space complexity is $O(ND + D^2) = O(D (N + D))$.

\subsection{SA} 
In the Simulated Annealing (SA) optimization algorithm, the complexity analysis considers several factors, including the number of iterations, the dimensionality of the problem space ($D$), the chosen cooling schedule, and the generation of neighbouring solutions.
The SA algorithm begins by randomly initializing a design pattern. During each iteration, neighbouring solutions are generated by applying perturbations or transformations to the current solution. We used random and swap mutation methods to generate neighbouring solutions in our implementation. The time complexity for finding neighbouring solutions using these methods is $O(D)$, where $D$ represents the dimensionality of the problem space. After generating a neighbouring solution, the algorithm evaluates its cost using a cost function. The evaluation of the cost function takes $O(f)$ time complexity. The algorithm then updates the best cost and best pattern by comparing them with the previous best pattern, which takes $O(D)$ time complexity.
The algorithm terminates after a maximum number of iterations, which serves as a termination criterion. Therefore, the overall time complexity of the SA algorithm can be expressed as $O({\tt maxIt} (2D + f))$, where ${\tt maxIt}$ denotes the maximum number of iterations. Additionally, the space complexity of the SA algorithm is $O(D)$ since it requires storing the current pattern and the best pattern during the optimization process.
\subsection{ALO}
In the Ant Lion Optimization (ALO) algorithm, time complexity varies according to the number of ants and ant lions, the roulette wheel selection method to pick the best antlions, dimensionality and the termination condition. 
The initial population of antlions and ants is randomly initialized in the search space. The time complexity of initialization is $O(ND)$, where $N$ is the number of antlions and $D$ is the dimensionality of the problem, and the space complexity is $O(2ND)$ for initializing the positions of both ants and antlions. Then, evaluate the cost of each ant lion in the population using the objective function and sort the ant lions based on their fitness values in descending order, which takes $O(Nf)$ time. The algorithm will create pits for each ant lion in each iteration based on its fitness value. The cost value determines the depth and size of the pit, where a lesser cost value corresponds to a deeper and larger pit. It allows each ant lion to perform a random walk within a certain radius around its current position. As one ant is prey to only one ant lion, a random walk of the ant is followed, and a roulette wheel selection method for selecting the antlion is used based on the antlion's fitness, which takes the worst case $O(N)$ time complexity.
For every iteration, the ants' random walk was observed. So, one antlion is selected using a roulette wheel operator for every ant because one ant is prey to only one antlion. An ant updates its position based on the random walk of selected antlion and elite antlion. This step takes $O(N^2 + ND)$. Now, calculate the cost of each ant based on its updated position and updates the antlion positions with its prey position, which takes $O(Nf)$. And then, calculate the best antlion position based on the fitness value and updates the elite or the best solution, which takes $O(N)$. The algorithm terminates after the termination condition reaches the total number of iterations (master). 
So, the total time taken by ALO is $O({\tt maxIt} (N^2 + ND + Nf + N))$. The overall time complexity is $O({\tt maxIt}(N(N + D + f + 1)))$  and space complexity is $O(2ND)$.

\nobibliography*
\bibliography{aaai24}

\end{document}